\documentclass[letterpaper, 10 pt, conference]{ieeeconf}  

\IEEEoverridecommandlockouts                              

\usepackage{times}
\usepackage{epsfig}
\usepackage{graphicx}
\usepackage{amsmath}

\usepackage{subcaption}
\DeclareCaptionSubType*[Alph]{table}
\DeclareCaptionLabelFormat{mystyle}{Table~\bothIfFirst{#1}{ ̃}#2}
\usepackage{array}
\newcolumntype{P}[1]{>{\centering\arraybackslash}p{#1}}

\usepackage{comment}

\title{\LARGE \bf
Tell me what this is: Few-Shot Incremental Object Learning by a Robot
}

\author{Ali Ayub$^{1}$ and Alan R. Wagner$^{2}$
\thanks{$^{1}$Department of Electrical Engineering,
        The Pennsylvania State University, State College, PA 16802, USA
        {\tt\small aja5755@psu.edu}}%
\thanks{$^{2}$Department of Aerospace Engineering, The Pennsylvania State University,
        State College, PA 16802, USA
        {\tt\small alan.r.wagner@psu.edu}}%
}

\begin{document}

\maketitle
\thispagestyle{empty}
\pagestyle{empty}

\begin{abstract}
\label{sec:Abstract}
For many applications, robots will need to be incrementally trained to recognize the specific objects needed for an application. This paper presents a practical system for incrementally training a robot to recognize different object categories using only a small set of visual examples provided by a human. The paper uses a recently developed state-of-the-art method for few-shot incremental learning of objects. After learning the object classes incrementally, the robot performs a table cleaning task organizing objects into categories specified by the human. We also demonstrate the system's ability to learn arrangements of objects and predict missing or incorrectly placed objects. Experimental evaluations demonstrate that our approach achieves nearly the same performance as a system trained with all examples at one time (batch training), which constitutes a theoretical upper bound.  
\end{abstract}
\section{INTRODUCTION}
\label{sec:introduction}
\noindent We seek to develop a practical system that would allow a novice human to teach a robot about different object categories incrementally using only a small set of visual examples provided by the person. Imagine, for example, a domestic robot tasked with locating and organizing household items. We would like for this robot to be trained on what items it should organize by its non-expert owner and recognize that the items to be organized might change over time. Although it could be possible to train the system on an enormous corpus of data containing a vast number of objects, hoping that all of the objects that the robot will one day be asked to organize will be in the dataset, this approach seems destined for failure. Ideally, the robot should be taught about important objects incrementally, and, because people will demand quick results, from only a few examples. This paper contributes a step towards such a system.  

Deep learning has achieved remarkable success recognizing objects~\cite{He_2016_CVPR}. Yet, to train a deep neural network data must be presented as a batch \cite{french19}. Attempts to create deep models that learn incrementally have been hindered by a problem known as \textit{catastrophic forgetting}~\cite{french19}. Catastrophic forgetting occurs when incrementally training a system to recognize new classes, the system forgets the previously learned classes and the overall classification accuracy decreases. 
To overcome this problem, some systems have been developed for incremental learning~\cite{Rebuffi_2017_CVPR,Castro_2018_ECCV}. An object incremental learner is used to classify objects but is trained on only a limited number of classes per increment. While training on a new increment, the training data for the previously learned classes is not available to the learner during the current increment. In order to test the accuracy of incremental learner, the classifier is tested on the complete test set of all the classes learned so far. Hence on a robot, the purpose of incremental learning is to allow the robot to be continuously trained over its operational lifetime. Most incremental learning methods~\cite{Rebuffi_2017_CVPR,Castro_2018_ECCV}, however, avoid catastrophic forgetting by storing a portion of the training data from the earlier learned classes which makes them impractical on robots which usually have a limited amount of on-board memory available. Furthermore, current methods for incremental learning require a large amount of training data and are thus not suitable for training by a non-expert, likely impatient, human owner. 

This paper builds from our prior research on incremental learning~\cite{Ayub20} by evaluating a practical application of few-shot incremental learning in which a robot is taught novel object classes incrementally using a small set of visual examples provided by a human. An $n$-shot incremental learner (where $n$ is usually 1,5, or 10) recognizes objects but is only trained on $n$ examples per class for $k$ classes per increment. We use our centroid-based concept learning (CBCL) approach proposed in~\cite{Ayub20,Ayub_scenes20} which achieves state-of-the-art performance on incremental learning and scene classification benchmarks and can be used for few-shot incremental learning of objects for a table cleaning task. First, the robot is taught classes of objects to clear from a few examples provided by a human incrementally. The robot then identifies objects belonging to the trained classes from a clutter of objects on the table and organizes the objects by type. We also demonstrate the system's ability to learn different object arrangements as semantic concepts. For example, a fork, plate and a spoon can be described as a dinner table setting. After learning the object classes incrementally from the examples provided by the human, the robot then uses the object's location and incremental classification to learn about the object-arrangements from a single example provided by the human. The robot can then recognize missing or wrong objects in different object arrangements.

The remainder of the paper is organized as follows: Section \ref{sec:related_work} reviews the related work including prior incremental learning approaches and robotic applications. Section \ref{sec:methodology} describes our complete architecture for few-shot incremental learning on a robot. Section \ref{sec:experiments} presents empirical evaluations of the system. Finally, Section \ref{sec:conclusion} offers conclusions and directions for future research. 
\section{RELATED WORK}
\label{sec:related_work}

\noindent The related work is divided into two categories: computer vision approaches to incremental learning and robotics applications of incremental learning for object classification.
\subsection{Incremental Learning for Computer Vision}
\noindent 
Early approaches to incremental learning use a fixed length feature extractor which takes images as input and generates a vector of image features. One of the early approaches for incremental learning used a Nearest Class Mean (NCM) classifier, which computes the mean of the feature vectors for the training images and uses the closest mean feature vector for classification of an unlabeled image~\cite{Mensink13}. Our approach is broadly related to this idea, yet, as will be discussed below, with many refined details. 

More recent incremental learning methods rely on storing a fraction of old class data when learning a set of new classes~\cite{Rebuffi_2017_CVPR,Castro_2018_ECCV}. iCaRL~\cite{Rebuffi_2017_CVPR} is one of the first methods proposed for incremental learning that combines knowledge distillation~\cite{Hinton15} and NCM (Nearest Class Mean) classifier for incremental learning. Knowledge distillation uses a distillation loss term that forces the labels of the training data of previously learned classes to remain the same when learning new classes. iCaRL uses the old class data while learning a representation for new classes and uses the NCM classifier for classification of the old and new classes. EEIL~\cite{Castro_2018_ECCV} improves iCaRL with an end-to-end learning approach. The main issue with these prior approaches is the need to store old class data which is not practical when the memory budget is limited. To avoid storage of real samples, some approaches use generative-memory and regenerate samples of old classes using GANs or autoencoders \cite{Ayub_ICML_20,Ostapenko_2019_CVPR,Ayub_NIPS_20}, however the performance of these approaches is generally inferior to approaches that store real images. Moreover, one issue with all these prior incremental learning approaches is that they require a large amount of training data. Hence using these methods for a few-shot incremental learning problem results in extremely poor accuracy. 

\subsection{Incremental Learning Applications in Robotics}
\noindent For many real world robotics applications, it will be impractical for a novice human teacher to provide hundreds of examples for each incremental class the robot needs to learn. Nevertheless, people will need to incrementally train their robots. Research has considered methods for few-shot learning ~\cite{Snell17} but these approaches have, for the most part, only been tested on image datasets like CIFAR-10~\cite{Krizhevsky09}. Most of these methods also require that the complete dataset, including all of the classes to be learned, is available as a single batch.

Incremental learning of novel object classes by a robot has been attempted in the past. Ude et al. uses handcrafted visual features to represent objects that are placed in a robot's hands to get different views of the objects \cite{ude08}. This approach, however, uses the feature vectors of all new and old classes for training, hence storing old training data and then further evaluates the success of their system on the training data. 
Valipour et al. presented a method to incrementally learn objects on a robot from interaction with a human \cite{Valipour17}. They train a portion of a pre-trained CNN when learning new classes of objects from examples provided by a human teacher. Turkoglu et al. offer another deep learning approach for learning objects incrementally using mobile robots \cite{Turkoglu18}. They also use prior class data when learning new classes which limits the usefulness of this approach for realistic applications. Another deep learning approach \cite{Denninger18} uses random forests to incrementally learn object classes but they do not test their approach on a real robot. Further, their approach requires the complete training set of each class and cannot be applied for few-shot incremental learning. An extension of \cite{Valipour17} is presented in \cite{Dehghan19} for incremental object learning for a robot using a combination of a pre-trained feature extractor and NCM classifier which has been shown to be significantly inferior to other incremental learning approaches \cite{Rebuffi_2017_CVPR,Ayub20}. 

The problems with these prior methods are: 1) That the incremental learning algorithms (\cite{Turkoglu18,Dehghan19}) are not compared to other state-of-the-art methods (\cite{Rebuffi_2017_CVPR,Castro_2018_ECCV}) or on benchmark datasets ; 2) Most of these approaches train the robot on an object class and then use the same object that was used for training in the test phase, which positively skews the results and limits real world applicability (\cite{Valipour17,Dehghan19}); 3) Most of the approaches incrementally learn a very small number of new object classes (10 or fewer) using a robot, which, because of the small number of classes, does not exhibit catastrophic forgetting \cite{Valipour17,Turkoglu18,Dehghan19} (as shown in Section \ref{sec:experiments}). The real challenge of incremental learning arises when learning a larger set of classes and when there is no overlap between the test and training sets. Moreover, the robotics community would benefit from an evaluation that provides quantitative metrics for comparison with future systems.

\begin{figure}[t]
\centering
\includegraphics[scale=0.4]{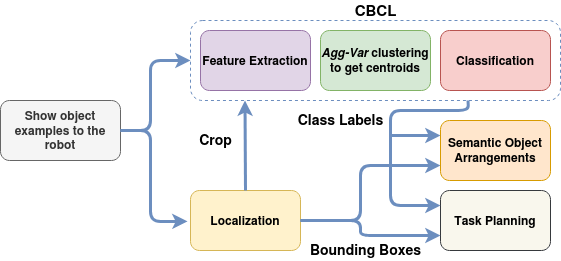}
\caption{\small Overall architecture of our approach. The robot learns new object classes incrementally using CBCL. The task planning module takes bounding boxes from the localization module and class labels from CBCL to move objects. The semantic object arrangements module also takes the bounding boxes and class labels to learn novel object arrangement concepts.}
\label{fig:few_shot_framework}
\end{figure}

\begin{figure}[t]
\centering
\includegraphics[scale=0.3]{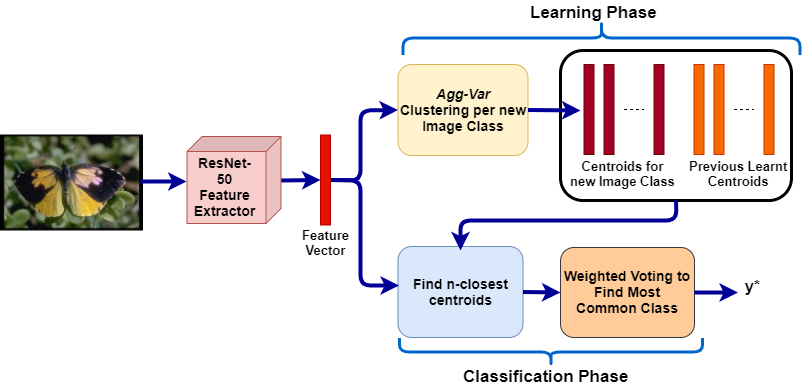}
\caption{\small For every new object class, the ResNet-50 feature extractor generates the CNN features of all the training images belonging to the object class and generates a set of centroids using \textit{Agg-Var} clustering algorithm, concatenates them with the centroids of previously learned classes and uses the complete set of centroids for classifying unlabeled test images}
\label{fig:framework}
\end{figure}
\section{INCREMENTAL LEARNING IN A FEW-SHOTS ON A ROBOT}
\label{sec:methodology}

\noindent Figure \ref{fig:few_shot_framework} presents the overall architecture for this work. The system can learn new object classes from only a few visual examples (5 or 10) provided by a human. The architecture consists of an incremental learning module which uses our recently developed algorithm called Centroid-Based Concept Learning (CBCL) \cite{Ayub20}, a localization module, task planning module, and semantic arrangements module.   

As input our system requests examples of object classes from a person. A simple text-based interaction module was designed to let the human teach the robot new object classes. When the person starts the teaching phase the robot waits for the person to show an example of the chosen object. The object is placed in front of the robot's hand camera and it captures an image of the object. Once images of an object have been collected, the CBCL algorithm is used to train a classifier to recognize the object. 

It should be noted that a human could have also taught the robot a new object by simply holding the object in their hand and allowing the robot to record different poses of the object. But this violates the spirit of few-shot learning since the video of an object may include hundreds or thousands of images to serve as training examples. Moreover, there may be some applications where the robot only has limited access to observe and learn about an object. Our approach also works with videos but we choose to evaluate the system as a true few-shot learning system. 

\subsection{Centroid-Based Concept Learning (CBCL)}
\label{sec:cbcl}

\noindent CBCL is a recently developed state-of-the-art method for incremental learning \cite{Ayub20}. CBCL has also been shown to learn objects from only a few examples whereas prior state-of-the-art methods require a large amount of training data, its memory footprint does not grow dramatically as it learns new classes, and its learning time is faster than the other, mostly deep learning, methods.

Figure~\ref{fig:framework} depicts the architecture for CBCL. It is composed of three modules: 1) a feature extractor, 2) \textit{Agg-Var} clustering, and 3) a weighted voting scheme for classification of unlabeled images. In the learning phase, once the human provides the robot with the training examples for a new class, the first step in CBCL is the generation of feature vectors from the images of the new class using a fixed feature extractor. In this paper, we use ResNet-50~\cite{He_2016_CVPR} pre-trained on ImageNet~\cite{Russakovsky15} as the feature extractor. 

Next, for each new image class $1 \leq y \leq N$, CBCL clusters all of the training images in the class provided by the human. \textit{Agg-Var} clustering begins by creating one centroid from the first image in the training set of class $y$. Next, for each image in the training set of the class, feature vector $x_i^y$ (for the $i$the image) is generated and compared using the euclidean distance to all the centroids for the class $y$. If the distance of $x_i^y$ to the closest centroid is below a pre-defined distance threshold $D$, the closest centroid is updated by calculating a weighted mean of the centroid and the feature vector $x_i^y$. If the distance between the $i$th image and the closest centroid is greater than the distance threshold $D$, a new centroid is created for class $y$ and equated to the feature vector $x_i^y$ of the $i$th image. 

The result of this process is a collection containing a set of centroids for the class $y$, $C^y = \{c_1^y, ..., c_{N^*_y}^y\}$, where $N^*_y$ is the number of centroids for class $y$. This process is applied to the sample set $X^y$ of each class incrementally once they become available to get a collection of centroids $C  = C^1, C^2, ..., C^N$ for all $N$ classes in a dataset. 
It should also be noted that \textit{Agg-Var} calculates the centroids for each class separately. Thus, CBCL's performance is not strongly impacted when the classes are presented incrementally. 

During the testing phase, to predict the label $y^*$ of a test image, CBCL uses the feature extractor to generate a feature vector $x$. Next, Euclidean distance is calculated between $x$ and the centroids of all the classes observed so far. Based on the calculated distances, the $n$ closest centroids to the unlabeled image are selected. The contribution of each of the $n$ closest centroids to the determination of the test image's class is a conditional summation:

\begin{equation}
    Pred(y) = \sum_{j=1}^{n} \frac{1}{dist(x,c_j)}[y_j=y]
\end{equation}

\noindent where $Pred(y)$ is the prediction weight of class $y$, $y_j$ is category label of $j$th closest centroid $c_j$ and $dist(x,c_j)$ is the Euclidean distance between $c_j$ and the feature vector $x$ of the test image. 
The test image is assigned the class label with the highest prediction weight.

\subsection{Object Localization}
\noindent This paper demonstrates few-shot incremental learning on a domestic cleaning task. For this task, multiple objects are present on a table in front of the robot. The robot must recognize and organize the objects by class. In order to perform this task the robot must first localize these objects. We use the RetinaNet \cite{Lin_2017_ICCV} for object localization. This network proposes image regions likely to contain objects. After passing the image through the RetinaNet, the bounding boxes identified by the network are sent to the incremental classification module (Subsection \ref{sec:cbcl}) for object recognition. The locations of bounding boxes are also passed on to the task planning (Subsection \ref{sec:task_planning_module}) and semantic object arrangements modules (Subsection \ref{sec:semantic_arrangement}). 

\subsection{Task Planning Module}
\label{sec:task_planning_module}
\noindent The task planning module takes the location of the object from the localization module and the class label of the object from CBCL. The module then moves the object classes specified by the human to a pre-defined location on the table. During testing, the robot has a cluttered mess of objects present on a table in front of it. These objects may belong to either a previously learned class or an unknown class. The human then asks the robot to move objects belonging to a particular class or multiple classes from the table. For example, "Clean all the toothbrushes from the table." The robot then uses the object localization and CBCL modules to localize and recognize the desired objects specified by the human. It then uses the suction gripper on its arm to pickup and move the desired objects from the table to the pre-defined location. 

\subsection{Semantic Object Arrangements Module} 
\label{sec:semantic_arrangement}
\noindent The robot can also learn object arrangements using a single training example provided by the human. When a new set of objects is presented, the robot utilizes the localization and classification modules to get the locations and labels of the objects and stores this information in the form of a centroid (different from the centroids of object classes generated by CBCL). After learning different object arrangements, the robot can then further predict the arrangement of a collection of objects. For example, in the test phase if a couple of objects are presented to the robot, it can use the  localization and classification modules to predict the locations and labels of the objects to create an object arrangement feature vector. Next, the closest object arrangement centroid to the test feature vector is used to identify the closest object arrangement centroid and to predict if an object is missing or wrong in the test arrangement.

\section{EXPERIMENTS}
\label{sec:experiments}
\noindent We evaluate our method for 5-shot and 10-shot incremental learning using the Baxter robot manufactured by Rethink robotics. In the sections below we present the implementation details for our experiments and evaluate our approach on 5-shot and 10-shot incremental learning by comparing it to other methods. We then demonstrate our method on a domestic cleaning application. Finally, we show that the robot can further learn higher level semantic concepts related to combination of objects and then predict about unknown situations. 

\begin{figure}[t]
\centering
\includegraphics[scale=0.5]{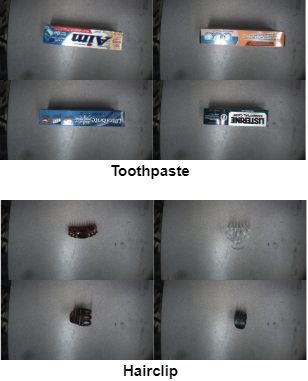}
\caption{\small Some example images of object classes toothpaste and hair clip learned by the robot using the camera in the robot's hand.}
\label{fig:objects}
\end{figure}

\begin{figure*}[t]
\centering
\includegraphics[scale=0.37]{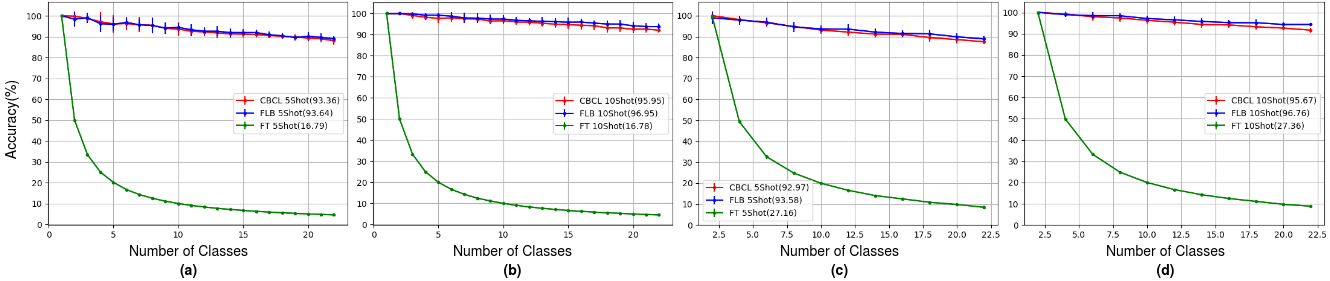}
\caption{\small Comparison of CBCL to the Few-shot Learning Baseline (upper bound) and Fine-Tuning (FT) for 5-shot and 10-shot incremental learning in terms of classification accuracy (\%). (a) and (b) show results for 5-shot and 10-shot incremental learning with 1 class per increment, while (c) and (d) show results for 5-shot and 10-shot incremental learning with 2 classes per increment. Mean and standard deviations of the classification accuracy is reported for each increment with 10 executions. Average incremental accuracy is shown in brackets for each method.}
\label{fig:few_shot_results}
\end{figure*}

\subsection{Implementation Details}
\noindent The Keras deep learning framework~\cite{chollet2015} was used to implement the neural network models. An Nvidia 1070 RTX GPU was used for all feature extraction and training. All of the input images were resized to $256 \times 256$ and randomly cropped to $224\times 224$ as the input to the network. ResNet-50 pre-trained on Imagenet was used as a feature extractor for CBCL and other methods. 

The first experiment compares our method for few-shot incremental learning to Few-shot Learning Baseline (FLB) and Fine-Tuning (FT) which were trained for 100 epochs in each increment using a fixed learning rate of 0.001 and cross-entropy loss with minibatches of size 8 optimized using stochastic gradient descent. For CBCL, for each batch of new classes, the hyper-parameters $D$ (distance threshold) and $n$ (number of closest centroids used for classification) were tuned using cross-validation. Only the previously learned centroids and the training data for the new classes were used for hyper-parameter tuning.

\begin{figure}[t]
\centering
\includegraphics[scale=0.35]{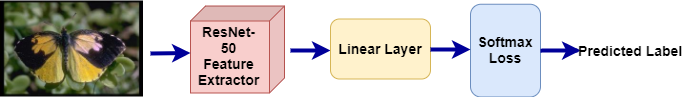}
\caption{\small Few-shot Learning Baseline (FLB) architecture}
\label{fig:baseline}
\end{figure}

\subsection{Few-shot Incremental Object Classification Experiment} 
\label{sec:few_shot_incremental}

\noindent This experiment evaluates the accuracy of CBCL on a robot and verifies the robot's ability to learn to recognize new objects incrementally using only a few examples (5 or 10) provided by a person. This experiment only uses the CBCL portion of the architecture in Figure \ref{fig:few_shot_framework}. Subsequent experiments utilize additional portions of the architecture.

A total of 22 object classes were used for this experiment. Common household items were chosen. Figure \ref{fig:objects} shows some examples of the object classes toothpaste and hair clip. For each object class, 15 different representative objects were purchased. Two different poses were used for each of the 15 objects per class, hence for each object class we had a total of 30 images. The robot was trained on one or two different randomly chosen classes per increment. During training, at each increment the robot was presented with 2 images of 15 objects for 2 classes; we also test with one class per increment. For 5-shot and 10-shot incremental learning settings, 5 and 10 images (out of the 30) were used as training images, respectively, and the rest were used as test images. 
As shown in Figure \ref{fig:objects}, the training and test images are realistic and not idealized. The background of the objects is not perfect since the table has several discolorations and some of the object sizes are rather small. Moreover, the lighting conditions are not ideal and some objects are transparent.

CBCL was compared to two other methods: Fine-Tuning (FT) and a Few-shot Learning Baseline (FLB). The features from the pre-trained ResNet-50 neural network were used for both methods. For both of the methods, ResNet features were passed to a linear layer which was trained using a softmax loss (Figure \ref{fig:baseline}). 
This procedure follows the prior few-shot learning research \cite{Chen19}. FT simply uses the linear layer trained on the previous classes and adapts it to the incoming classes from the new increment. FLB trains the final linear layer using softmax loss on the complete training set of all the new and old classes in each increment. In other words, FLB does not learn incrementally. Rather, it represents the upper bound for the classification accuracy at each increment. This experiment was conducted over ten rounds in which the ordering of the classes presented to the robot during each increment was random. The selection of the training and test images for each class was also randomized. Average accuracies over all ten rounds are presented as results.

Figure \ref{fig:few_shot_results} compares CBCL against the two methods for the 5-shot and 10-shot incremental learning experiments with 1 and 2 classes per increment. For the first increment there is no catastophic forgetting so all the methods perform the same. For the rest of the increments, CBCL outperforms FT by significant margins and the difference in classification accuracy between the two methods increases as the number of classes increase for 5-shot and 10-shot cases with both 1 and 2 classes per increment. For FT, catastrophic forgetting (described in the introduction) causes the dramatic decrease in accuracy. Furthermore, as expected, catastrophic forgetting for FT is greater when learning 1 class per increment than when learning 2 classes per increment. 

FLB, as a theoretical upperbound, generates the best results at each increment for all cases because it uses the training data of all the classes. In terms of average incremental accuracy, CBCL is only \textbf{0.28\%} and \textbf{1.0\%} lower than the FLB (the theoretical upper bound) for 5-shot and 10-shot incremental learning with one class per increment. For 2 classes per increment, for the first 5 increments CBCL and FLB have the same performance for both 5-shot and 10-shot incremental learning (Figure \ref{fig:few_shot_results} (c) and (d)). For the next 6 increments CBCL's accuracy is only slightly lower than the baseline (within \textbf{1\%}) for both 5-shot and 10-shot incremental learning. In terms of average incremental accuracy, CBCL is only \textbf{0.61\%} and \textbf{1.09\%} lower than FLB for 5-shot and 10-shot incremental learning experiments, respectively. 

The results indicate that our method's performance is almost as good as a neural network baseline trained with all of the classes' training data in one batch. CBCL's performance is closer to the FLB upperbound when using one class per increment. Moreover, CBCL's performance is also closer to the upperbound for 5-shot incremental learning. This may suggest that CBCL is best suited for incremental learning situations where data is scarce. 

During the teaching phase, the time required to collect a single object image by the robot takes only $\sim2$ seconds and the time required to extract features for the image takes about 10 seconds. In the learning phase, the time required to learn a set of centroids for a new batch of classes takes around 1 second. During prediction phase, time required to extract the features and predict the label of a test image takes 11 seconds (10 seconds for feature extraction and 1 second for prediction).

\begin{figure}[t]
\centering
\includegraphics[scale=0.4]{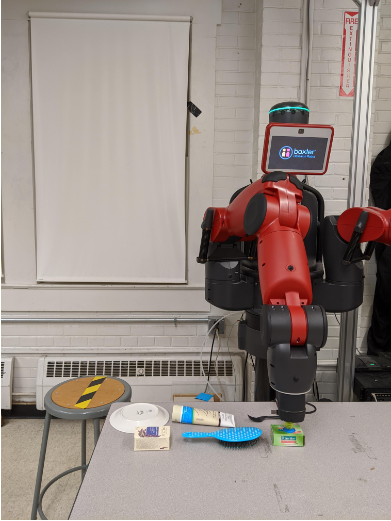}
\caption{\small An example of Baxter robot cleaning objects belonging to class \textit{soap} from a set of objects in front of it's arm camera on the table.}
\label{fig:baxter}
\end{figure}

\subsection{Table Cleaning Experiment}
\noindent A robot cleaning task is presented as a practical demonstration of our method for few-shot incremental learning approach. For this experiment, a robot is taught new object classes in a few-shot incremental learning setting using the architecture in Figure \ref{fig:few_shot_framework}. This experiment uses the CBCL, localization, and task planning modules. After teaching the robot a set of objects in each increment we asked it to clean a specific object class from a clutter of objects on the table. The objects on the table were from any class, already learned or unknown to the robot. As depicted in the architecture the first step is for the robot to localize the objects. Next the robot crops the bounding boxes of the objects and pass them to CBCL for classification. After finding the object labels, it then finds the objects on the table that belong to the desired class to be cleaned. The robot then uses the suction gripper on the end effector to pick and move the desired objects to a pre-defined collection location on the table. Suction points on objects were pre-defined. Figure \ref{fig:baxter} shows an example of a table cleaning task with different objects present on the table in front of the Baxter robot. 

This experiment was performed a total of 10 times after learning all 22 object classes incrementally with 2 classes per increment in a 5-shot incremental learning setting. In each of the 10 runs of this experiment we asked the robot to move different object classes. For each run, there are a total of 6 objects present on the table in front of the robot with two objects belonging to the object class to be moved. 

Three different kinds of errors can cause the robot to fail in completing this task. The first is a detection error in which the object localization module fails to detect and draw a bounding box around the desired object. The second is the classification error in which the robot fails to predict the correct class of the detected object. The last type of error is a movement error in which the object is correctly detected and recognized, yet the robot fails to pick and move it to the desired place. 

\begin{table}[t]
\centering
\small
\begin{tabular}{ |P{3cm}|P{1.5cm}| }
     \hline
    \textbf{Error Type} & \textbf{Error (\%)} \\
    \hline
    Detection Error & 20 \\
    \hline
    Classification Error & 12.5 \\
    \hline
    Movement Error & 0 \\
 \hline
 \end{tabular}
 \caption{Three different kinds of errors encountered during the table cleaning task by the robot}
 \label{tab:task_planning_error}
 \end{table}

Table \ref{tab:task_planning_error} reports the average values of these three errors for the 10 runs of the experiment. The highest error rate (detection errors) was 20\%. The classification error rate was in accordance with the prior results reported in Figure \ref{fig:few_shot_results}(c). Since, the suction gripper was used to pick and place objects, the robot picked and moved all the objects successfully.  

For this experiment, the object localization module takes approximately 30 seconds to find the bounding boxes of the objects in the image, classification of each of the bounding boxes takes about 11 seconds when all 22 classes are learned. Finally, moving each object from the table to a pre-defined place requires approximately 10 seconds. The total time necessary to identify and clean an object is 51 seconds.

\begin{figure}[t]
\centering
\includegraphics[scale=0.15]{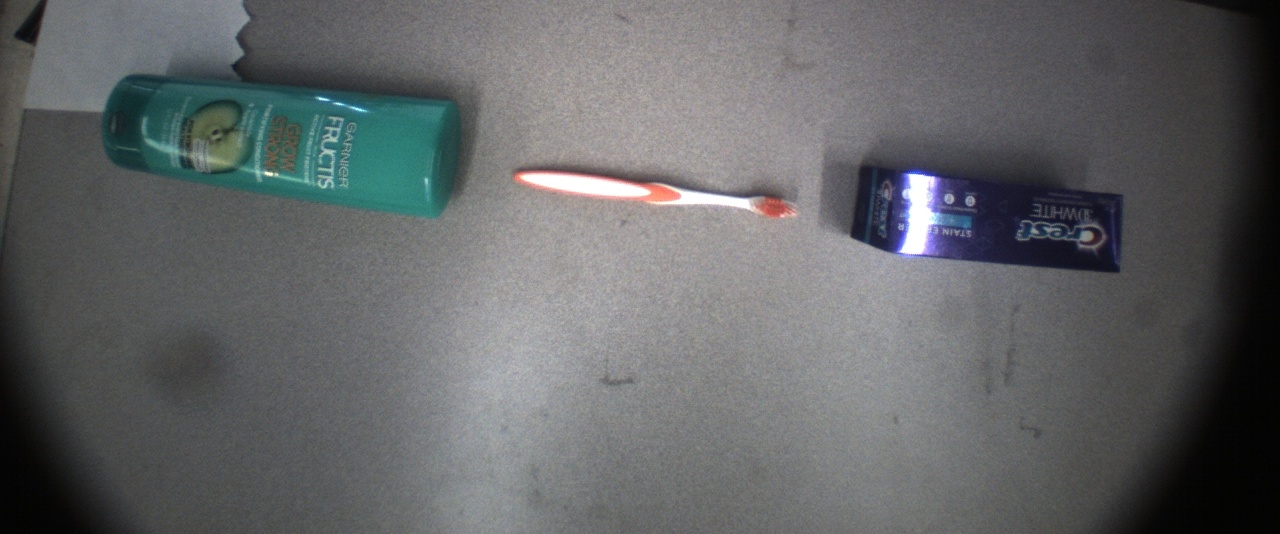}
\caption{\small A training example of arrangement of objects found in the bathroom (shampoo, toothbrush, toothpaste) shown to the robot.}
\label{fig:object_arrangement}
\end{figure}

\subsection{Predicting Missing or Wrong Objects Experiment}
\noindent As a final experiment the ability of the robot to learn and correct different arrangements of objects was evaluated. This experiment uses all of the modules from the architecture presented in Figure \ref{fig:few_shot_framework} (except task planning module).  A total of 5 different object arrangements were taught each involving 3 different objects. Figure \ref{fig:object_arrangement} shows an example of an arrangement of objects used to train the robot. Each arrangement was represented as a vector containing object labels and spatial relationships between the objects. Only four spatial relations (Left, Right, Above, Below) between objects were considered. The first 22 elements of the feature vector represent the 22 object classes learned by the robot. The presence of an object is represented as 1 in the feature vector, 0 if the object is absent. The spatial relation left and right are represented by a single matrix of size 22$\times$22, in which a 1 at a location $i,j$ represents that object of class $i$ is on the left of object of class $j$. The above and below relations are represented by a similarly constructed  22$\times$22 matrix. Both of these matrices are flattened and appended to the complete feature vector making the size of a single feature vector to be $22+22\times22+22\times22=990$. 

For each object arrangement, the robot was presented with a single example. It used the localization and classification modules (5-shot) to get the object locations and labels. The robot stored these relations in the form of 990$\times$1 feature vector (centroid) mentioned above. In the test phase the robot was presented with 20 different object arrangements, 10 of which contained a missing object and the other 10 contained an incorrect object in comparison with the 5 object arrangements learned by the robot. During the test phase, we assumed that the newly presented arrangements were a type of a previously learned arrangement. Hence, by comparing the test arrangement with the earlier learned arrangements it was possible to recognize incorrect arrangements.

The object examples used in the test samples were different from the ones in the training images. For each test sample, the robot first created a 990$\times$1 feature vector and then found the closest centroid using Euclidean distance. Using the closest centroid, the robot then predicted which of the objects were either missing or wrong in the test feature vector. In case of a wrong object, the robot further predicted the correct object in its place using the closest centroid. It should be noted that there can be two centroids at an equal distance from the test feature vector, in that case the robot predicts two possible objects that can replace the missing or wrong object in the test feature vector.

For each of the 20 test scenarios, there was only one object that was missing or wrong but because of classification and detection errors there were cases when the robot had to make predictions about more than one object as missing or wrong. Of the 20 cases, 9 cases required the robot to make two predictions and 11 cases required only one prediction. In these cases where the robot must predict two objects rather than one, the accuracy was (5/9$\times$100) 56\%. For cases in which there was no detection error and the robot only had to predict one object the accuracy was 100\%. The overall accuracy for all 20 test cases was 16/20$\times$100=80\%.

\section{CONCLUSION}
\label{sec:conclusion}
\noindent This paper presents and evaluates a method for few-shot incremental learning on a robot. We have shown that our approach allows a robot to learn different categories of objects from only a few examples. Moreover, even after learning 22 different categories of objects our approach correctly recognizes previously learned objects at a rate of approximately 90\% accuracy on a robot. We have further demonstrated this work on a practical application such as domestic cleaning of a table. Our final experiment shows that after learning the objects incrementally, the robot learns different semantic object arrangement concepts from a single example provided by the human and then uses these concepts to predict the missing or wrong objects in different object arrangement settings.

This research is not without its assumptions and limitations. For example, we assume a table-top environment in which the robot knows how and where to pick and place objects on the table. Moreover, object labels were provided by the experimenter rather than naive human subjects. Finally, the arrangement of objects was limited to a small set of fixed patterns.   
Nevertheless, we believe that this paper makes several important contributions. First, a robot that incrementally learns may be capable of dynamically learning task or situation specific objects, allowing for more general robotics applications. Second, learning from only a few-shots may be necessary in order to work with non-expert humans which have limited patience for teaching a robot. Third, we provide quantitative technical evaluations of few-shot incremental learning for a robot. Similar future systems can evaluate the quality of their proposed methods against the work presented here. We intend to release the images of the 22 object classes as a dataset for a fair comparison. Finally, this work may evolve into new applications and competencies for existing robots, such as domestic cleaning and industrial packaging.





\section*{ACKNOWLEDGMENT}
\noindent This work was supported by Air Force Office of Scientific Research contract FA9550-17-1-0017.


{\small
\bibliographystyle{IEEEtran.bst}
\bibliography{main}

\begin{thebibliography}{10}
\providecommand{\url}[1]{#1}
\csname url@rmstyle\endcsname
\providecommand{\newblock}{\relax}
\providecommand{\bibinfo}[2]{#2}
\providecommand\BIBentrySTDinterwordspacing{\spaceskip=0pt\relax}
\providecommand\BIBentryALTinterwordstretchfactor{4}
\providecommand\BIBentryALTinterwordspacing{\spaceskip=\fontdimen2\font plus
\BIBentryALTinterwordstretchfactor\fontdimen3\font minus
  \fontdimen4\font\relax}
\providecommand\BIBforeignlanguage[2]{{%
\expandafter\ifx\csname l@#1\endcsname\relax
\typeout{** WARNING: IEEEtran.bst: No hyphenation pattern has been}%
\typeout{** loaded for the language `#1'. Using the pattern for}%
\typeout{** the default language instead.}%
\else
\language=\csname l@#1\endcsname
\fi
#2}}

\bibitem{He_2016_CVPR}
K.~He, X.~Zhang, S.~Ren, and J.~Sun, ``Deep residual learning for image
  recognition,'' in \emph{The IEEE Conference on Computer Vision and Pattern
  Recognition (CVPR)}, June 2016.

\bibitem{french19}
R.~M. French, ``Dynamically constraining connectionist networks to produce
  distributed, orthogonal representations to reduce catastrophic
  interference,'' \emph{Proceedings of the Sixteenth Annual Conference of the
  Cognitive Science Society}, pp. 335--340, 2019.

\bibitem{Rebuffi_2017_CVPR}
S.-A. Rebuffi, A.~Kolesnikov, G.~Sperl, and C.~H. Lampert, ``i{C}a{RL}:
  Incremental classifier and representation learning,'' in \emph{The IEEE
  Conference on Computer Vision and Pattern Recognition (CVPR)}, July 2017.

\bibitem{Castro_2018_ECCV}
F.~M. Castro, M.~J. Marin-Jimenez, N.~Guil, C.~Schmid, and K.~Alahari,
  ``End-to-end incremental learning,'' in \emph{The European Conference on
  Computer Vision (ECCV)}, September 2018.

\bibitem{Ayub20}
A.~Ayub and A.~R. Wagner, ``Cognitively-inspired model for incremental learning
  using a few examples,'' in \emph{The IEEE/CVF Conference on Computer Vision
  and Pattern Recognition (CVPR) Workshops}, June 2020.

\bibitem{Ayub_scenes20}
A.~Ayub and A.~Wagner, ``{CBCL}: {B}rain inspired model for {RGB-D} indoor
  scene classification,'' \emph{arXiv:1911.00155}, 2019.

\bibitem{Mensink13}
T.~Mensink, J.~Verbeek, F.~Perronnin, and G.~Csurka, ``Distance-based image
  classification: Generalizing to new classes at near-zero cost,'' \emph{IEEE
  Transactions on Pattern Analysis and Machine Intelligence}, vol.~35, no.~11,
  pp. 2624--2637, Nov 2013.

\bibitem{Hinton15}
G.~Hinton, O.~Vinyals, and J.~Dean, ``Distilling the knowledge in a neural
  network,'' in \emph{NIPS Deep Learning and Representation Learning Workshop},
  2015.

\bibitem{Ayub_ICML_20}
A.~Ayub and A.~R. Wagner, ``Storing encoded episodes as concepts for continual
  learning,'' \emph{arXiv:2007.06637}, 2020.

\bibitem{Ostapenko_2019_CVPR}
O.~Ostapenko, M.~Puscas, T.~Klein, P.~Jahnichen, and M.~Nabi, ``Learning to
  remember: A synaptic plasticity driven framework for continual learning,'' in
  \emph{The IEEE Conference on Computer Vision and Pattern Recognition (CVPR)},
  June 2019, pp. 11\,321--11\,329.

\bibitem{Ayub_NIPS_20}
A.~Ayub and A.~R. Wagner, ``{EEC}: {L}earning to encode and regenerate images
  for continual learning,'' 2020, under Review, NeurIPS.

\bibitem{Snell17}
J.~Snell, K.~Swersky, and R.~Zemel, ``Prototypical networks for few-shot
  learning,'' in \emph{Advances in Neural Information Processing Systems 30},
  I.~Guyon, U.~V. Luxburg, S.~Bengio, H.~Wallach, R.~Fergus, S.~Vishwanathan,
  and R.~Garnett, Eds.\hskip 1em plus 0.5em minus 0.4em\relax Curran
  Associates, Inc., 2017, pp. 4077--4087.

\bibitem{Krizhevsky09}
A.~Krizhevsky, ``Learning multiple layers of features from tiny images,'' 2009,
  technical report, University of Toronto.

\bibitem{ude08}
A.~Ude, D.~Omrčen, and G.~Cheng, ``Making object learning and recognition an
  active process,'' \emph{International Journal of Humanoid Robotics}, vol.~05,
  no.~02, p. 267–286, 2008.

\bibitem{Valipour17}
S.~Valipour, C.~P. Quintero, and M.~J{\"a}gersand, ``Incremental learning for
  robot perception through hri,'' \emph{2017 IEEE/RSJ International Conference
  on Intelligent Robots and Systems (IROS)}, pp. 2772--2777, 2017.

\bibitem{Turkoglu18}
M.~O. Turkoglu, F.~B.~T. Haar, and N.~van~der Stap, ``Incremental
  learning-based adaptive object recognition for mobile robots,'' in \emph{2018
  IEEE/RSJ International Conference on Intelligent Robots and Systems (IROS)},
  Oct 2018, pp. 6263--6268.

\bibitem{Denninger18}
M.~Denninger and R.~Triebel, ``Persistent anytime learning of objects from
  unseen classes,'' in \emph{2018 IEEE/RSJ International Conference on
  Intelligent Robots and Systems (IROS)}, Oct 2018, pp. 4075--4082.

\bibitem{Dehghan19}
M.~Dehghan, Z.~Zhang, M.~Siam, J.~Jin, L.~Petrich, and M.~Jagersand, ``Online
  object and task learning via human robot interaction,'' in \emph{2019
  International Conference on Robotics and Automation (ICRA)}, May 2019, pp.
  2132--2138.

\bibitem{Russakovsky15}
O.~Russakovsky, J.~Deng, H.~Su, J.~Krause, S.~Satheesh, S.~Ma, Z.~Huang,
  A.~Karpathy, A.~Khosla, M.~Bernstein, A.~C. Berg, and L.~Fei-Fei, ``Imagenet
  large scale visual recognition challenge,'' \emph{Int. J. Comput. Vision},
  vol. 115, no.~3, pp. 211--252, Dec. 2015.

\bibitem{Lin_2017_ICCV}
T.-Y. Lin, P.~Goyal, R.~Girshick, K.~He, and P.~Dollar, ``Focal loss for dense
  object detection,'' in \emph{The IEEE International Conference on Computer
  Vision (ICCV)}, Oct 2017.

\bibitem{chollet2015}
F.~Chollet \emph{et~al.}, ``Keras,'' \url{https://github.com/fchollet/keras},
  2015.

\bibitem{Chen19}
W.-Y. Chen, Y.-C. Liu, Z.~Kira, Y.-C.~F. Wang, and J.-B. Huang, ``A closer look
  at few-shot classification,'' in \emph{International Conference on Learning
  Representations}, 2019.

\end{thebibliography}
}

\end{document}